\documentclass[letterpaper,conference]{IEEEtran}
\IEEEoverridecommandlockouts

\usepackage{amsmath,amssymb,amsfonts}
\usepackage{algorithmic}
\usepackage{algorithm}
\usepackage{array}
\usepackage[caption=false,font=normalsize,labelfont=sf,textfont=sf]{subfig}
\usepackage{textcomp}
\usepackage{stfloats}
\usepackage{url}
\usepackage{verbatim}
\usepackage{graphicx}
\usepackage{cite}
\usepackage{pifont}
\usepackage{hyperref}
\usepackage{multirow}
\usepackage{balance}

\begin{document}

\title{Towards Mechatronics Approach of System Design, Verification and Validation for Autonomous Vehicles}

\author{Chinmay Samak$^{\ast}$, Tanmay Samak$^{\ast}$, Venkat Krovi
\thanks{$^{\ast}$These authors contributed equally.}
\thanks{C. V. Samak, T. V. Samak and V. N. Krovi are with the Automation, Robotics and Mechatronics Laboratory (ARMLab), Department of Automotive Engineering, Clemson University International Center for Automotive Research (CU-ICAR), Greenville, SC 29607, USA. Email: {\tt\small {\{\href{mailto:csamak@clemson.edu}{csamak}, \href{mailto:tsamak@clemson.edu}{tsamak}, \href{mailto:vkrovi@clemson.edu}{vkrovi}\}@clemson.edu}}}}

\maketitle


\begin{abstract}
Modern-day autonomous vehicles are increasingly becoming complex multidisciplinary systems composed of mechanical, electrical, electronic, computing and information sub-systems. Furthermore, the individual constituent technologies employed for developing autonomous vehicles have started maturing up to a point, where it seems beneficial to start looking at the synergistic integration of these components into sub-systems, systems, and potentially, system-of-systems. Hence, this work applies the principles of mechatronics approach of system design, verification and validation for the development of autonomous vehicles. Particularly, we discuss leveraging multidisciplinary co-design practices along with virtual, hybrid and physical prototyping and testing within a concurrent engineering framework to develop and validate a scaled autonomous vehicle using the AutoDRIVE Ecosystem. We also describe a case-study of autonomous parking application using a modular probabilistic framework to illustrate the benefits of the proposed approach.
\end{abstract}



\section{Introduction}
\label{Section: Introduction}

\IEEEPARstart{A}{utomotive} vehicles have evolved significantly over the course of time \cite{Beiker2016}. The gradual transition from purely mechanical automobiles to those with greater incorporation of electrical, electronic and computer-controlled sub-systems occurred in phases over the course of the past century; with each phase improving performance, convenience and reliability of these systems. Modern vehicles are increasingly adopting electrical, electronic, computing and information sub-systems along with software algorithms for low-level control as well as high-level advanced driver assistance system (ADAS) or autonomous driving (AD) features \cite{Meissner2020}. This naturally brings in the interplay between different levels of mechanical, electrical, electronic, networking and software sub-systems among a single vehicle system, thereby transforming them from purely mechanical systems, which they were in the past, to complex multidisciplinary systems \cite{Gumiel2022}. As such, while it may have been justifiable for earlier ADAS/AD feature developers to focus on core software development, the increasing complexity and interdisciplinary nature of modern automotive systems can benefit from synergistic hardware-software co-design complemented with integrated verification and validation by following the mechatronics principles.

\begin{figure}[t]
	\centering
	\includegraphics[width=\linewidth]{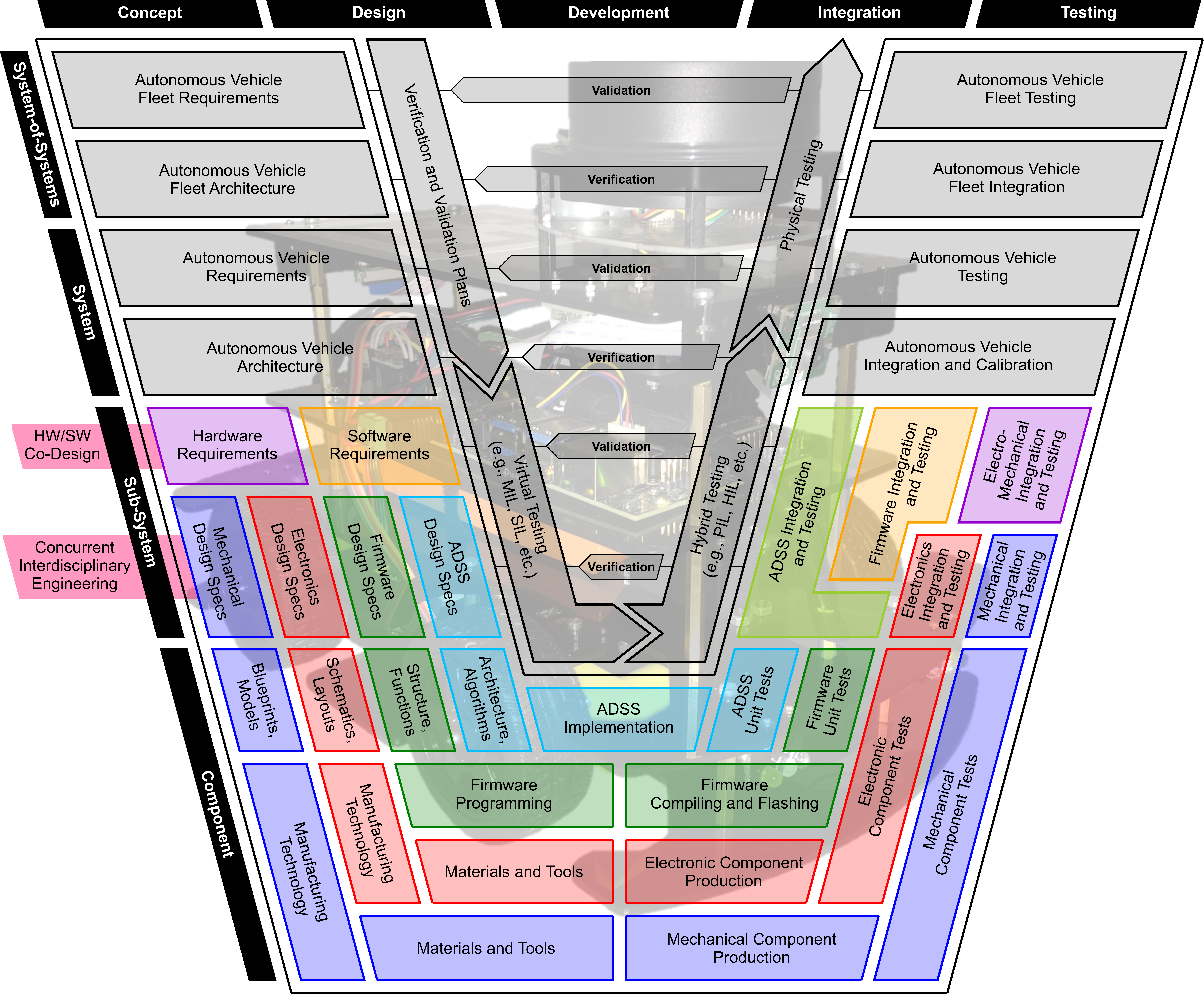}
	\caption{Extended V-model fostering mechatronics approach of system design, verification and validation for autonomous vehicles. The model depicts evolution of a concept into a product through decomposition, design, development, integration and testing across component, sub-system, system and system-of-systems levels in a unified concurrent interdisciplinary engineering framework.}
	\label{fig1}
\end{figure}

Mechatronics engineering \cite{Bolton1995, deSilva2004, Onwubolu2005} focuses on concurrent and synergistic integration of mechanical, electrical and electronics engineering, computer science and information technology for development and validation of complex interdisciplinary systems. This ideology is derived from the fact that various components of a ``mechatronic'' system, often belonging to a multitude of disciplines, influence each other and hence have a design impact at the component, sub-system, system and system-of-systems levels. The resulting ``mechatronic'' realization now builds on capabilities endowed by the various constituent layers. In such a milieu, the system development approach has also evolved from relatively ad-hoc to the more formal V-model \cite{Gausemeier2002}, building on the modular software development and validation roadmap \cite{Brohl1995}. This model has evolved through several progressions \cite{Eigner2017} and our work seeks to further formalize the adoption of mechatronics approach of system conceptualization, design, development, integration and testing for autonomous vehicles (refer Fig. \ref{fig1}).

\begin{figure*}[t]
	\centering
	\includegraphics[width=\linewidth]{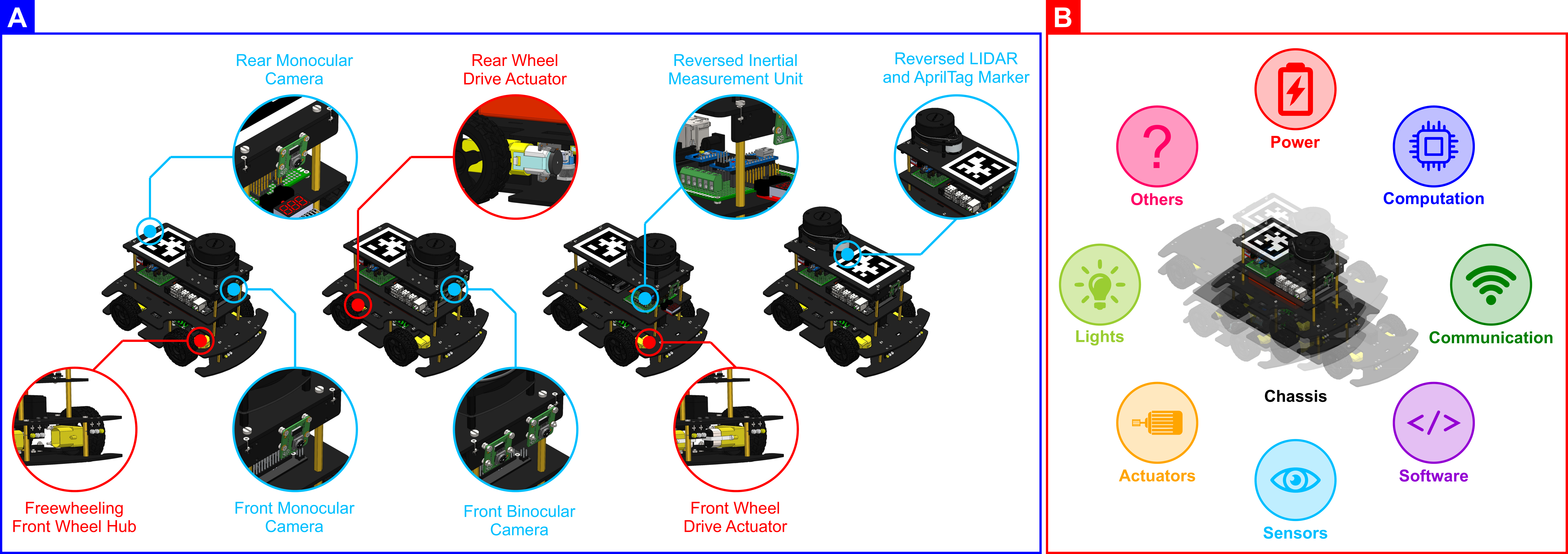}
	\caption{AutoDRIVE Ecosystem fosters mechatronics design principles at two levels: [A] primitive reconfigurability allows permutations and combinations of addition, removal or replacement of selective components and sub-assemblies of the vehicle to better suit the application; [B] advanced reconfigurability allows complete modification of existing hardware and software architectures, and provides an opportunity for introducing new features and functionalities to the ecosystem.}
	\label{fig2}
\end{figure*}

A recent book\cite{Pathrose2022} highlights best practices for industrial design, development and validation of autonomous vehicles and notes the significant adoption of model-based design (MBD) for system integration and testing. However, similar adoption of such streamlined workflows by academia has lagged behind \cite{Pathrose2022}. This gap could be explained by the virtue of standardization (e.g. ISO 26262 \cite{ISO26262}, ISO/IEC 33061 \cite{ISO33061}, VDI 2221 \cite{VDI2221}, VDI 2206 \cite{VDI2206}, AUTOSAR \cite{Furst2009}, etc.) in industries versus the fact that majority of academic projects are deployed using fragmented hardware-software ecosystems (e.g. hobby platforms) with a key focus on developing low-cost initial proof-of-concept implementations. Additionally, such an opportunistic and potentially uninformed selection of hardware \cite{MITRacecar2017, F1TENTH2019, MuSHR2019} and software \cite{Gazebo, CARLA, Cognata} toolchains hinders adoption of co-design and concurrent engineering thinking to full extent.

In this paper, we discuss the design philosophy and one of the key motivation factors behind AutoDRIVE Ecosystem\footnote{Website: \texttt{\url{https://autodrive-ecosystem.github.io}}} \cite{AutoDRIVEEcosystem2022, AutoDRIVEReport2021} – a tightly integrated ecosystem for developing, simulating and deploying autonomous vehicles across scales and complexities. The said philosophy involves adopting and promoting mechatronics approach of system design, verification and validation for autonomous vehicles, with an emphasis on creating a streamlined pathway for transitioning towards the ultimate industrial practice. This paper also describes a detailed case-study, which demonstrates the methodical adoption of mechatronics approach for designing, developing and validating a scaled vehicle in the context of autonomous parking\footnote{Video: \texttt{\url{https://youtu.be/piCyvTM2dek}}} application using a modular probabilistic framework.


\section{Multidisciplinary Design}
\label{Section: Multidisciplinary Design}

AutoDRIVE offers an open-access, open-interface and flexible ecosystem for autonomous vehicle development by permitting access to and alteration of hardware as well as software aspects of the multidisciplinary system design. Particularly, this ecosystem and its various modules  can assist in executing each step of every phase and level in the proposed V-model (refer Fig. \ref{fig1}), thereby making it an apt framework for demonstrating the claims and contributions of this work.

AutoDRIVE Ecosystem offers the following two levels of design reconfigurability, thereby promoting hardware-software co-design (refer Fig. \ref{fig2}).

\begin{itemize}
	\item \textbf{Primitive Reconfigurability:} AutoDRIVE's native vehicle, called ``Nigel'', is modular enough to support out-of-the-box hardware reconfigurability in terms of adding, removing or replacing selective components and sub-assemblies of the vehicle for perception (encoders, IPS, IMU, cameras, LIDAR, etc.), computation (high/low-level), communication (wired/wireless), actuation (Ackermann/skid-steered, front/rear/all-wheel drive), illumination (head/tail-lights, turn/reverse indicators) and power (battery, power electronics) in addition to flexibly updating the vehicle firmware and/or autonomous driving software stack (ADSS) to better suit target application(s).
	\item \textbf{Advanced Reconfigurability:} The completely open-hardware, open-software architecture of AutoDRIVE Ecosystem allows modification of vehicle chassis parameters (different form factors and component mounting profiles), powertrain configuration (variable driving performance), as well as firmware and ADSS architecture (software flexibility).
\end{itemize}

The fundamental step in system design is requirement specification, without which the design cannot be truly validated to be right or wrong, it can only be surprising \cite{Young1985}. Since AutoDRIVE was intended to be a generic ecosystem for rapidly prototyping autonomous driving solutions, the requirement elicitation resulted in a superset of requirements demanded by the application case study discussed in this paper. Furthermore, being an open-access ecosystem, AutoDRIVE allows for updating the designs of various components, sub-systems and systems for expanding the ecosystem. That being said, following is a summary of functional requirement specifications for Nigel as of this version of the ecosystem.

\begin{figure*}[t]
	\centering
	\includegraphics[width=\linewidth]{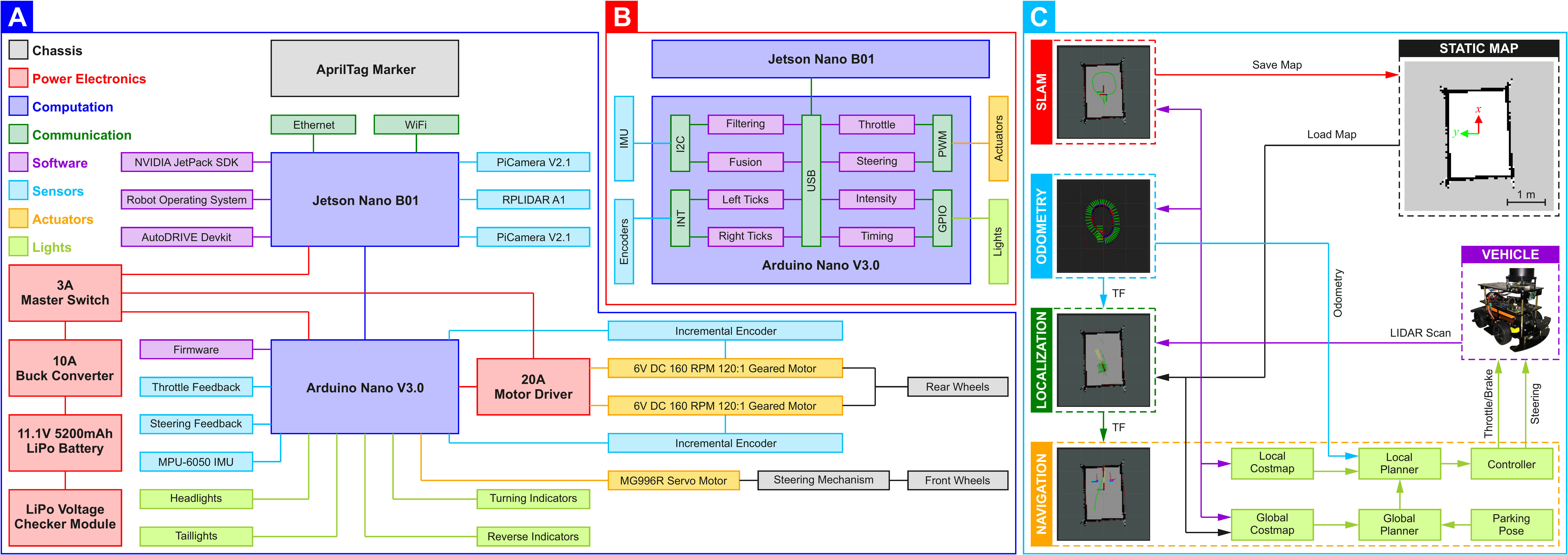}
	\caption{Conceptualization and design of scaled autonomous vehicle: [A] hardware-software architecture; [B] firmware design specifications; [C] modular perception, planning and control architecture for autonomous parking application.}
	\label{fig3}
\end{figure*}

\begin{itemize}
	\item General design guidelines:
	\begin{itemize}
		\item{Open-source\footnote{GitHub: \texttt{\url{https://github.com/AutoDRIVE-Ecosystem}}} hardware and software}
		\item{Inexpensive ($\leq$ \$500) and user-friendly architecture}
		\item{Manufacturing technology agnostic (e.g. 3D printing, laser cutting, etc.)  designs}
		\item{Modularly reconfigurable components/sub-systems}
		\item{Integrated and comprehensive resources and tools}
		
	\end{itemize}
	\item Perception sub-system shall offer:
	\begin{itemize}
		\item{Ranging measurements (360$^\circ$ FOV, $\leq$ 1$^\circ$ resolution)}
		\item{RGB visual feed (preferably front as well as rear)}
		\item{Positional coordinates $\{X, Y, Z\}$ (error $\leq$ 5e-2 m)}
		\item{Inertial measurements $\{a_x, a_y, a_z, \omega_x, \omega_y, \omega_z\}$ and AHRS estimates $\{\phi_x, \theta_y, \psi_z\}$}
		\item{Actuation feedback measurements $\{\tau, \delta\}$}
	\end{itemize}
	\item Computation and communication sub-systems shall offer:
	\begin{itemize}
		\item{Hierarchical computation topology (high/low-level)}
		\item{GPU-enabled high-level edge computation platform}
		\item{Embedded low-level computation platform}
		\item{Vehicle-to-everything (V2X) communication}
	\end{itemize}
	\item Locomotion and signaling sub-systems shall offer:
	\begin{itemize}
		\item{Kinodynamically constrained drivetrain and steering}
		\item{Standard automotive lighting and signaling}
	\end{itemize}
\end{itemize}

The functional system requirements were decomposed into mechanical, electronics, firmware and ADSS design specifications and carefully studied to analyze any potential trade-offs so as to finalize the components and ultimately come up with a refined system architecture design (refer Fig. \ref{fig3}).

The proposed hardware-software architecture of the scaled autonomous vehicle system is divided into eight sub-systems viz. chassis, power, computation, communication, software, sensors, actuators and lights, each with its own share of components (refer Fig. \hyperref[fig3]{\ref*{fig3}-A}). The embedded firmware architecture for low-level data acquisition and control is depicted in Fig. \hyperref[fig3]{\ref*{fig3}-B}, which links the data sources to the respective sinks after processing the information. Finally, Fig. \hyperref[fig3]{\ref*{fig3}-C} depicts high-level architecture of the autonomous parking solution described in this paper. Particularly, it is shown how this candidate autonomy solution uses modular algorithms for simultaneous localization and mapping (SLAM) \cite{HectorSLAM2011}, odometry estimation \cite{RF2O2016}, localization \cite{AMCL2001}, global \cite{AStar1968} and local \cite{TEBPlanner2017} path planning, and motion control. Implementation descriptions are necessarily brief due to the space limitations; however, further details can be found in this technical report \cite{AutoDRIVEReport2021}.


\section{Virtual Prototyping and Testing}
\label{Section: Virtual Prototyping and Testing}

Virtual prototypes help expedite the design process by validating the designs against system requirements through simulation, and suggesting design revisions at an early stage. Additionally, they help reduce testing costs, improve safety and facilitate collaboration among design teams. Furthermore, virtual prototypes may be the only viable option for conducting tests such as corner and edge case analysis, variability and repeatability analysis, robustness and stress testing in high-risk conditions, to name a few.

\begin{figure*}[t]
	\centering
	\includegraphics[width=\linewidth]{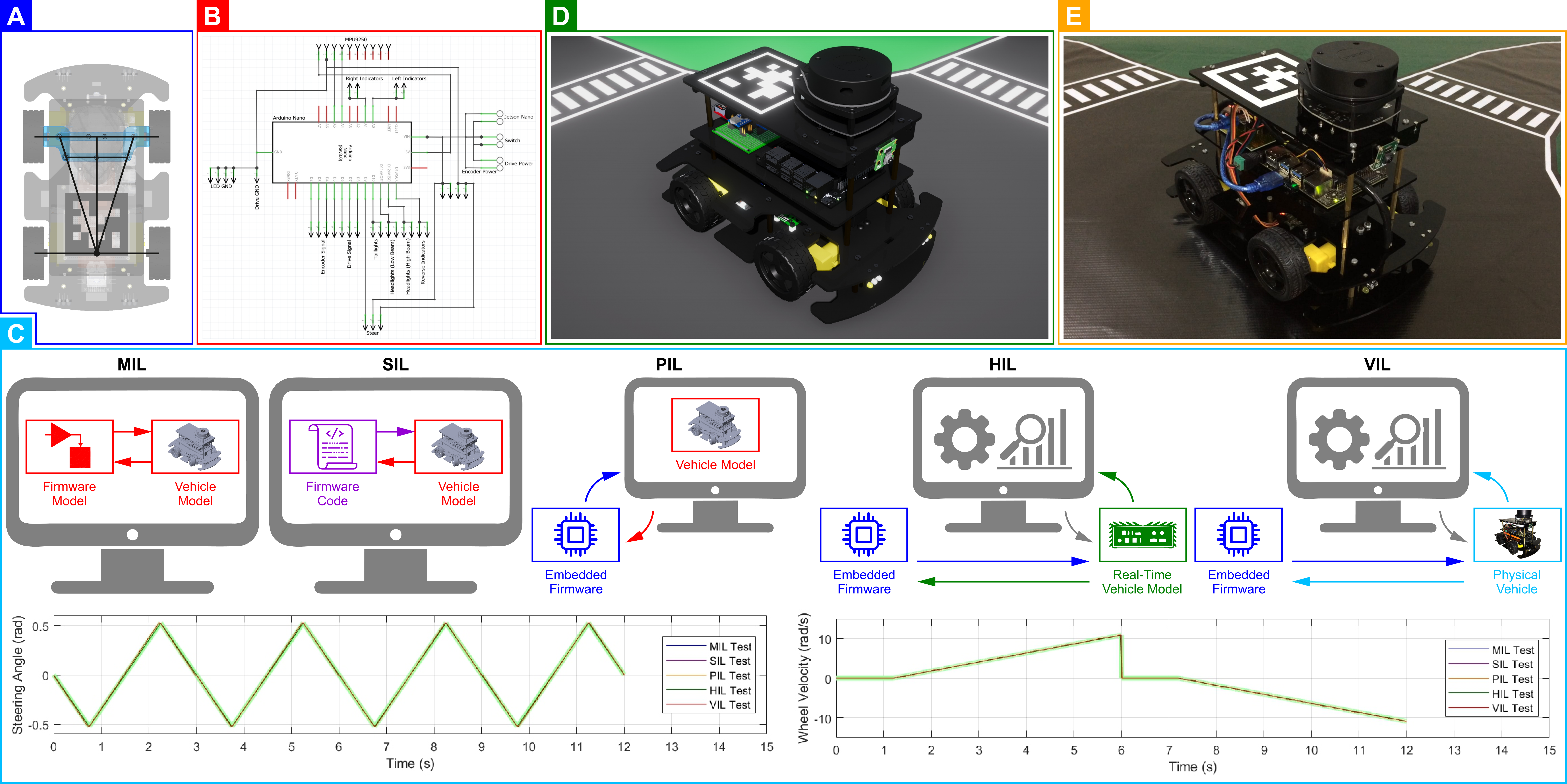}
	\caption{Development and system integration of scaled autonomous vehicle: [A] mechanical assembly; [B] electronic schematic; [C] MBD workflow depicting MIL, SIL, PIL, HIL and VIL testing of vehicle firmware; [D] virtual prototype in AutoDRIVE Simulator; [E] physical prototype in AutoDRIVE Testbed.}
	\label{fig4}
\end{figure*}

The scaled autonomous vehicle system was virtually prototyped and tested in three phases. First, the mechanical specifications, motions and fit were carefully analyzed using a parametric computer aided design (CAD) assembly of the system in conjunction with the physical modeling approach for multi-body dynamic systems (refer Fig. \hyperref[fig4]{\ref*{fig4}-A}). Parallelly, the electronic sub-systems were prototyped using the physical modeling approach, and also by adopting electronic design automation (EDA) workflow (refer Fig. \hyperref[fig4]{\ref*{fig4}-B}). Next, the firmware for low-level control (front wheel steering angle and rear wheel velocity) of the vehicle was verified to produce reliable results (within a specified tolerance of 3e-2 rad for steering angle and 3e-1 rad/s for wheel velocity) through model-in-the-loop (MIL) and software-in-the-loop (SIL) testing (refer Fig. \hyperref[fig4]{\ref*{fig4}-C}).

The knowledge gained through this process was used to update the AutoDRIVE Simulator (refer Fig. \hyperref[fig4]{\ref*{fig4}-D}) from its initial version discussed in \cite{AutoDRIVESimulator2021, AutoDRIVESimulatorReport2020} to the one described in \cite{AutoDRIVEEcosystem2022, AutoDRIVEReport2021}. The updated simulator was then employed for verification and validation of individual ADSS modules and finally, the integrated autonomous parking solution was also verified using the same toolchain (refer Fig. \hyperref[fig5]{\ref*{fig5}-A}). Particularly, we tested the vehicle in multiple environments, which included unit tests for validating the SLAM, odometry, localization, planning and control algorithms, followed by verification of the integrated pipeline with and without the addition of dynamic obstacles, which were absent while mapping the environment. The autonomous navigation behavior was analyzed for 5 trials and verified to fit within an acceptable tolerance threshold of 2.5e-2 m; the acceptable parking pose tolerance was set to be 5e-2 m for linear positions in X and Y directions and 8.73e-2 rad for the angular orientation about Z-axis.

\section{Hybrid Prototyping and Testing}
\label{Section: Hybrid Prototyping and Testing}

All models or virtual prototypes involve certain degrees of abstraction, ranging from model fidelity to simulation settings, and as such, cannot be treated as perfect representations of their real-world counterparts. Therefore, once the virtual prototyping and preliminary testing of the system has been accomplished, the next step is to prototype and validate it in a hybrid fashion (partly virtual and partly physical), focusing more on high-level system integration. This method of hybrid prototyping and testing is extremely beneficial since it follows a gradual transition from simulation to reality, thereby enabling a more faithful system verification framework and providing a room for potential design revisions even before complete physical prototyping is accomplished.

The scaled vehicle system was subjected to hybrid testing by running processor-in-the-loop (PIL), hardware-in-the-loop (HIL) and vehicle-in-the-loop (VIL) tests on the embedded firmware for confirming minimum deviation from MIL and SIL results, specified by the same tolerance values of 3e-2 rad for steering angle and 3e-1 rad/s for wheel velocity (refer Fig. \hyperref[fig4]{\ref*{fig4}-C}). The performance of integrated autonomous vehicle system was then validated using hybrid testing in two phases.

First, we deployed the ADSS on the physical vehicle's on-board computer, which was interfaced with AutoDRIVE Simulator to receive live sensor feed from the virtual vehicle, process it and generate appropriate control commands, and finally relay these commands back to the simulated vehicle. Specifically, for the autonomous parking solution (refer Fig. \hyperref[fig5]{\ref*{fig5}-A}), we deployed and tested each of the SLAM, odometry, localization, planning and control algorithms for satisfactory performance. This was naturally followed by deployment and validation of the integrated pipeline for accomplishing reliable (within a precision tolerance of 2.5e-2 m) source-to-goal navigation (within a goal pose tolerance of 5e-2 m and 8.73e-2 rad) in different environments, wherein a subset of cases included dynamic obstacles as discussed earlier.

Next, we collected real-world sensor data using AutoDRIVE Testbed and replayed it as a real-time stimulus to the ADSS deployed on the physical vehicle's on-board computer running in-the-loop with AutoDRIVE Simulator. This way, we increased the ``real-world'' component of the hybrid test and verified the autonomous parking solution for expected performance (within same tolerance values as mentioned earlier). Particularly, the real-world data being collected/replayed was occupancy-grid map of the environment built by executing the SLAM module on the physical vehicle, which inherently resulted as a unit test of this module in real-world conditions. The simulated vehicle had to then localize against this real-world map while driving in the virtual scene and navigate autonomously from source to goal (parking) pose, which further tested the robustness of the integrated pipeline against minor environmental variations and/or vehicle behavior.


\section{Physical Prototyping and Testing}
\label{Section: Physical Prototyping and Testing}

\begin{figure*}[t]
	\centering
	\includegraphics[width=\linewidth]{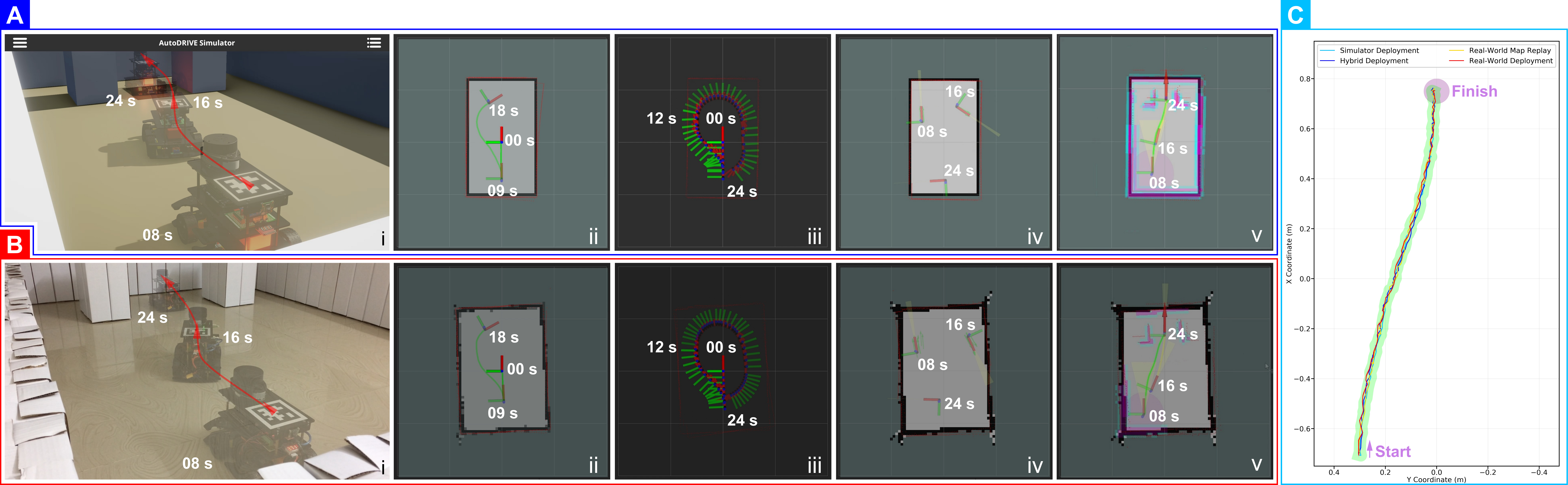}
	\caption{Verification and validation of scaled autonomous vehicle performance: [A] virtual/hybrid and [B] physical validation of (i) integrated system, unit testing of (ii) SLAM, (iii) odometry, (iv) localization, (v) planning and control modules in AutoDRIVE Simulator/Testbed; [C] repeatability/reliability analysis represented as mean and standard deviation of 5 trials for each deployment type with acceptable trajectory tolerance in green and parking tolerance in purple.}
	\label{fig5}
\end{figure*}

Verification and validation of safety-critical systems such as autonomous vehicles is quite difficult owing to their inherent complexity. Although virtual and hybrid testing approaches expedite the initial design and development process, these systems need to be ultimately tested in proving-ground or real-world conditions to validate their performance. Additionally, the question of simulation fidelity and benchmarking of virtual prototypes against their physical counterparts must be addressed. Consequently, once these systems confirm satisfactory performance under hybrid testing conditions, the next and final stage in mechatronic system development is physical prototyping and testing (refer Fig. \hyperref[fig4]{\ref*{fig4}-E}).

In order to physically validate the autonomous vehicle (refer Fig. \hyperref[fig5]{\ref*{fig5}-B}), we initially carried out unit tests to confirm the performance of each of the individulal algorithms for SLAM, odometry, localization, planning and control, followed by deployment of the integrated software stack for autonomous parking application (refer Fig. \hyperref[fig5]{\ref*{fig5}-C}). The vehicle was confirmed to exhibit a reliable (within a precision tolerance of 2.5e-2 m) source-to-goal navigation (within a goal pose tolerance of 5e-2 m and 8.73e-2 rad). Once again, as in case of virtual prototyping, the robustness of ADSS was tested by introducing dynamic obstacles that were not present while environment mapping was performed. This physical experimentation also helped benchmark AutoDRIVE Simulator against AutoDRIVE Testbed; that is to say, it was confirmed that the virtual and physical prototypes exploited for validating the aforementioned candidate autonomy algorithm have high resemblance in terms of physical (kinematics and dynamics) and graphical (dimensional accuracy and photorealistic rendering) fidelity of the vehicle as well as environment.


\section{Conclusion}
\label{Section: Conclusion}

The key contributions of this work include: (i) formal extension of the V-model describing mechatronics approach of system design, verification and validation for autonomous vehicles; (ii) adoption of AutoDRIVE Ecosystem for modular and reconfigurable multidisciplinary design of autonomous vehicles across scales; and (iii) setting up a streamlined workflow for virtual, hybrid as well as physical prototyping and testing of autonomous vehicles in the context of an autonomous parking application case study.

Particularly, we presented an extended V-model fostering mechatronics approach of system design, verification and validation for autonomous vehicles. Further, we discussed how AutoDRIVE Ecosystem aims to adopt and promote the mechatronics approach for autonomous vehicle development across scales and inculcate a habit of following it from academic education and research to industrial deployments. We also demonstrated the methodical adoption of mechatronics approach for designing, developing and validating a scaled autonomous vehicle in the context of a detailed case study pertaining to autonomous parking using a modular probabilistic framework; including both qualitative and quantitative remarks. We showed that the design, development as well as verification and validation of the scaled autonomous vehicle with regard to the aforementioned case study could be successfully accomplished within a stringent time-frame of about one month \cite{AutoDRIVEReport2021}. It is to be noted that although the exact timeline of any multidisciplinary project may vary depending upon factors such as skill set, experience and number of individuals involved, lead time in the supply chain, etc., the mechatronics approach definitely proves to be efficient in terms of minimizing the design-development iterations by the virtue of synergistic integration in a concurrent engineering thinking framework. This provides a room for the rectification of any design issues early in the development cycle, thereby increasing the chances of successful verification and validation with minimal loss of time and resources.

In order to better compare the mechatronics approach against others (in the context of autonomous vehicles), we propose some qualitative (degree of integration, flexibility, functionality, safety, reliability, etc.) and quantitative (development time, cost, performance, efficiency, complexity, etc.) metrics, which we plan to investigate in a future work.


\balance
\bibliographystyle{IEEEtran}
\bibliography{IEEEabrv,References}

\end{document}